%% file: main.tex
\documentclass{article}
\usepackage{spconf,amsmath,graphicx,comment,algorithm,algorithmic,subcaption}
\usepackage{tipa}
\usepackage[table,xcdraw]{xcolor}
\usepackage{multirow}
\usepackage{todonotes}
\usepackage{booktabs}
\usepackage{colortbl}
\usepackage{hyperref}
\usepackage{siunitx}
\usepackage{makecell}
\usepackage{rotating}
\usepackage{paralist}
\def\vx{{\mathbf x}}
\def\vy{{\mathbf{y}}}

\newcommand{\sysname}{\textsc{AdaptLMD}}
\newcommand{\prnnt}{P_\text{rnnt}}
\newcommand{\scoreindividual}{\Tilde{S}_{\text{disc}}}
\newcommand{\scoretotal}{\Tilde{S}_{\text{tot}}}
\newcommand{\ILM}{P_\text{ILM}}
\newcommand{\ILMroll}{P_\text{roll}}
\newcommand{\AM}{P_\text{IAM}}
\newcommand{\ID}{\mathbf{0}}
\newcommand{\vocab}{\mathcal{V}}
\newcommand{\blankT}{\epsilon}

\newcommand{\K}{D_\text{adapt}}

\title{Adaptive Discounting of Implicit Language Models in RNN-Transducers}
\name{Vinit Unni$^{1,\dagger}$, Shreya Khare$^{2,\dagger}$, Ashish Mittal$^{1,2}$, Preethi Jyothi$^1$, Sunita Sarawagi$^1$, Samarth Bharadwaj$^2$\thanks{$^{\dagger}$ Equal contribution.}}
\address{$^1$Indian Institute of Technology Bombay, India\\ $^2$IBM Research, India}
\begin{document}
%
\maketitle
\begin{abstract}
RNN-Transducer (RNN-T) models have become synonymous with streaming end-to-end ASR systems. While they perform competitively on a number of evaluation categories, rare words pose a serious challenge to RNN-T models. One main reason for the degradation in performance on rare words is that the language model (LM) internal to RNN-Ts can become overconfident and lead to hallucinated predictions that are acoustically inconsistent with the underlying speech. To address this issue, we propose a lightweight adaptive LM discounting technique \sysname, that can be used with any RNN-T architecture without requiring any external resources or additional parameters. \sysname\ uses a two-pronged approach: \begin{inparaenum}
\item Randomly mask the prediction network output to encourage the RNN-T to not be overly reliant on it's outputs.
\item Dynamically choose when to discount the implicit LM (ILM) based on rarity of recently predicted tokens and divergence between ILM and implicit acoustic model (IAM)  scores.
\end{inparaenum}
Comparing \sysname\ to a competitive RNN-T baseline, we obtain up to $4\%$ and $14\%$ relative reductions in overall WER and rare word PER, respectively, on a conversational, code-mixed Hindi-English ASR task.
\end{abstract}
%
\input{Intro}
\input{relatedwork}

\section{Background: RNN-Transducers}
\label{sec:RNNT}
Consider an acoustic input $\vx = \{ x_1, x_2, x_3 ... x_T \}$ with corresponding output text transcript  $\vy = \{ y_1, y_2, y_3 ... y_K \}$ where each $y_j \in \vocab$, the output vocabulary. $\vocab$ also includes a special blank token $\blankT$.
An RNN-T model comprises a Prediction Network (PN), a Transcription Network (TN) and a joint network (Figure \ref{fig:rnnt}).
The PN takes as input the previously emitted \emph{non-blank} tokens $\vy_{<u}= y_1\ldots y_{u-1}$, and auto-regressively generates an encoding for the next output token as $g_u$. 
The TN takes as input the acoustic signal $\vx$ and outputs encoder states for each of the time steps $\mathbf{h} = \{ h_1, h_2, h_3, ... h_T \}$.  Henceforth, we will also use implicit acoustic model (IAM) and implicit LM (ILM) to refer to the TN and PN, respectively.
The joint network $J$ for each time step $t \le T$ and token position $u \le K$ takes as inputs the encoded vectors $h_t$ and $g_u$ and generates a probability distribution over the vocabulary $\vocab$ as:
\begin{equation}
\label{eq:prnnt}
\begin{aligned}
    \prnnt(y_{t,u} | \vy_{<u}, x_t) &=& \text{softmax}\{J(h_t \oplus g_u)\} \\
    g_u = \text{PN}(\vy_{<u}), &~~~& h_t = \text{TN}(\vx, t)
\end{aligned}
\end{equation}
Traditionally, the $\oplus$ is an addition operation, however, other operations can be used too. $J$ is any feed-forward network, PN is typically an RNN and TN a Transformer. The output probability $\prnnt$ over the $[t,u]$ space are marginalized during training to maximize the likelihood of the token sequence $y_1,\ldots,y_K$ using an efficient dynamic programming algorithm.  During decoding, beam-search is used to find the best possible $[t,u]$ alignment and output token sequence~\cite{Graves2012}.

 \begin{figure}
\centering
     \begin{subfigure}[b]{\columnwidth}
         \centering
             \includegraphics[width=0.9\columnwidth]{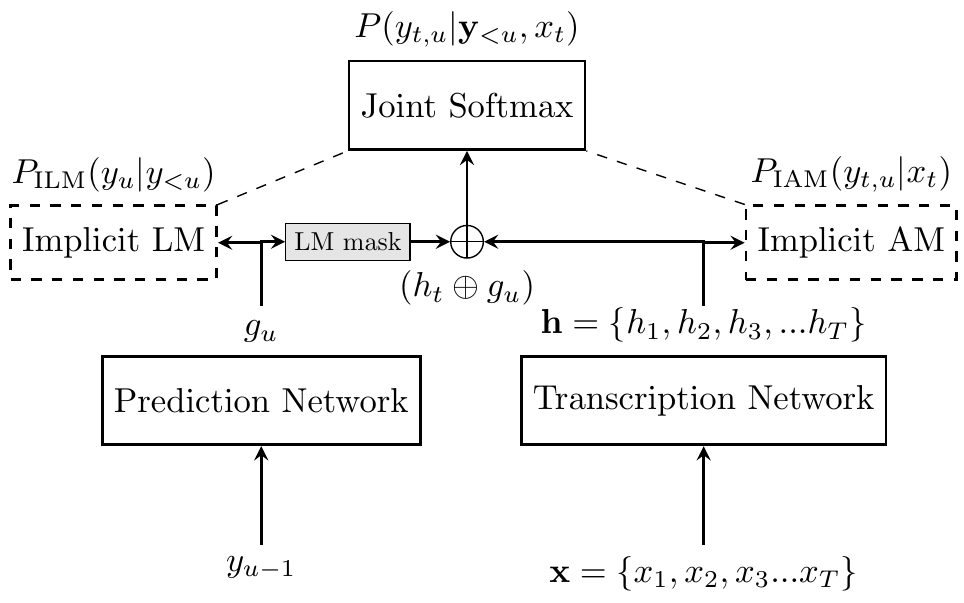}
         \caption*{(a) Training: LM masking and implicit LM+AM loss}
         \label{fig:discounting}
     \end{subfigure}%

     \begin{subfigure}[b]{\columnwidth}
         \centering
             \includegraphics[width=1\columnwidth]{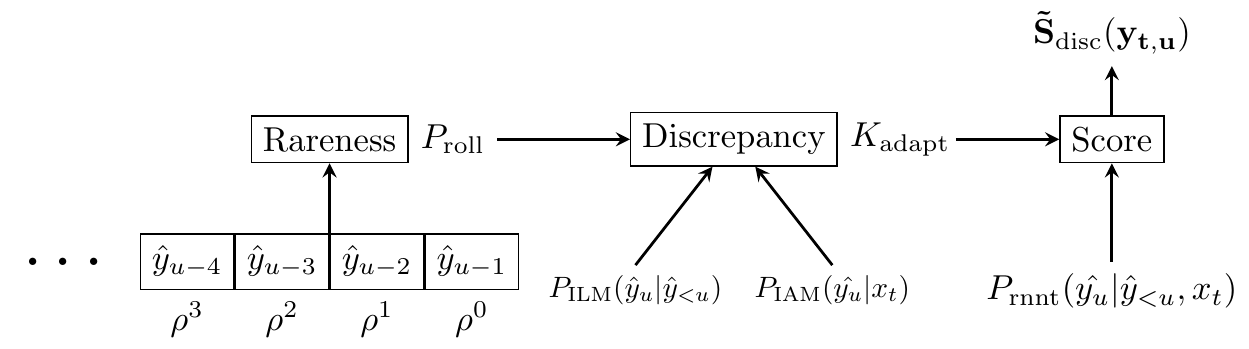}
         \caption*{(b) Decoding: Adaptive discounting of implicit LM}
         \label{fig:discounting}
     \end{subfigure}
     \caption{Overview of \sysname}
     \label{fig:rnnt}
     \label{fig:rnnt}
 \end{figure}


\section{Our Approach}
A limitation of the symmetric integration of the language encoding ($g_u$) with the acoustic encoding ($h_t$) is a systematic bias against rare words, often resulting in hallucination of words that have no connection with the underlying speech. We seek to temper such overt   influence of the ILM when we expect its predictions to be noisy.  Our main challenge is how to detect, during online decoding, if the ILM is misleading. 
We propose a light-weight method called \sysname\ that can be applied on existing RNN-T architectures without any external lexicons or additional parameters. 

\sysname\ differs from existing RNN-T based ASR in two ways.  First, during training we impose random masking of the ILM output $g_u$, so that the joint network learns to be robust to spurious embeddings by the ILM.
Second, during decoding we impose an {\em adaptive} discounting of the ILM on the output distribution $\prnnt$.  We dynamically choose when to discount based on two factors: 
(1) The discrepancy between the IAM and ILM signals, and 
    (2) The rarity of recent tokens in the current hypothesis.
%
%
We compute these terms using existing RNN-T modules and describe these in Sections~\ref{subsec:discr} and \ref{subsec:rare} respectively. Our formula for adaptive ILM discounting based on these two scores, and the overall decoding algorithm is described in Section~\ref{subsec:implicitLM}. The final training algorithm of \sysname\ appears in Section~\ref{subsec:train}.

\subsection{Discrepancy between LM  and AM}
\label{subsec:discr}
The RNN-T architecture allows easy extraction of the independent token distribution of the TN and PN components  by masking the $h_t$ and $g_u$, respectively.  We call these $\ILM$ and $\AM$ and define as:
\begin{align}
\label{eq:iam}
     \AM(y_{t}|\vx,t) &=& \text{softmax}\{J( h_t \oplus \ID)\} \\ \ILM(y_{t}|\vy_{<u}) &=& \text{softmax}\{J( \ID \oplus g_u)\}
\end{align}
where $\ID$ denotes an identity vector.
These distributions are trained along with the primary $\prnnt$ as elaborated in Sec~\ref{subsec:train}. The distance between $\ILM,\AM$ can be any divergence metric. We use \emph{KL-divergence}: \begin{equation}\text{D}(\ILM || \AM)=\sum_{y \in \vocab} \ILM(y|\vy_{<u})\log\frac{\ILM(y|\vy_{<u})}{\AM(y|x_t)} \end{equation}. 
\noindent 
\newcommand{\hyp}{\vy}
\subsection{Rarity of Recent Tokens}
\label{subsec:rare}
We  enable discounting at rare locations within a currently decoded hypothesis. Instead of external resources we depend on the implicit LM $\ILM$ to characterize rare regions.  Let the current hypothesis be $\hyp = \{ y_1, y_2, y_3, ... y_{u-1} \}$. We define a quantity $\ILMroll(\hyp)$ that indicates a rolling average of the ILM probabilities over the recent tokens in the hypothesis $\hyp$.  We choose  a $0 \leq \rho \leq 1$ to compute $\ILMroll(\hyp+y_{u})$ of $\hyp$ extended by a new token $y_{u}$ 
at step $u$ incrementally from the $\ILMroll(\hyp)$ as:
\begin{equation}
\label{eq:ilmroll}
    \begin{gathered}
     \ILMroll(\hyp+y_u) 
     = \rho \ILMroll(\hyp) + \ILM(y_u) 
 \end{gathered}
\end{equation}
As we want the discounting to be higher for rarer words where the implicit LM confidence on the hypothesis generated so far is low,
we use $1-\ILMroll(\vy)$ as a measure of the local rarity of recent tokens of a current hypothesis $\vy$.



%

\subsection{Decoding with Implicit LM Discounting}
\label{subsec:implicitLM}
Using the above two scores, we adaptively discount the influence of the ILM in regions where the it is less confident and where there is discrepancy between the ILM and IAM as follows:
Let $\hyp=y_1,\ldots y_{u-1}$ denote the current hypothesis on the beam.  The discounted score of the next token $y_{t,u}$ by which we extend $\hyp$ at position $u$ from frame $t$ is then computed as follows:
%
\begin{align*}
\label{eq:discounting}
     \scoreindividual(y_{t,u}| & \hyp,x_t) = \log \prnnt(y_{t,u}|\hyp, x_t) \\  
      &-\lambda \max(0,\K(y_{t,u},\hyp)) \log \ILM(y_u|\hyp) \\ \nonumber
     \K(y,\vy)=&
     \begin{cases}
        (1-\ILMroll(\hyp))\text{D}(\ILM(y) || \AM(y))&\text{if } y\neq\blankT\\
        0  &  \text{else.}
     \end{cases} \nonumber
\end{align*}
\noindent where $\lambda$ is a tunable parameter. 
While decoding we use the above discounted scores $\scoreindividual$ as a drop-in replacement for the original $\log\prnnt$ scores in the beam-search algorithm. With this discounted score, we defer to predictions of the implicit AM and discount the predictions by the $\ILM$ in regions of low confidence (indicated by $\ILMroll$). The only additional state we need to maintain for each hypothesis on the beam is $\ILMroll$.  The rest of the inference algorithm stays unchanged.   We extended the \emph{Alignment-Length Asynchronous Decoding (ALSD)}~\cite{9053040} with the above changes as the decoding algorithm for \sysname. \\

\subsection{Overall Training Objective}
\label{subsec:train}
Our training strategy differs from RNN-T's in two ways: (1) We use random masking of the PN encoding to encourage the RNN-T not to be overly reliant on its output, and (2) we introduce extra losses for training the $\ILM$ and $\AM$.
\begin{equation}
\label{eq:train_loss}
\begin{gathered}
   \log \sum_{\text{align t}}\prnnt(y_{t,u}|h_t\oplus \text{maskedLM}_{\eta}(g_u)) \\
   +\alpha  \log \sum_{\text{align t}}\ILM(y_u|\ID \oplus g_u) \\
   + \beta\log \sum_{\text{align t}}\AM(y_{t,u}|h_t \oplus \ID) 
\end{gathered}
\end{equation}
In eqn~\eqref{eq:train_loss}, $\text{maskedLM}_{\eta}(g_u)$ denotes the random masking of the TN where with a probability  $\eta$ we replace $g_u$ with $\ID$.
\section{Experiments and Results}
\input{results.tex}
\section{Conclusion}
We propose \sysname, a new adaptive LM discounting technique that is dynamically invoked and can be used within any RNN-T model without requiring any additional resources. This technique particularly benefits rare words for which the LM internal to an RNN-T can generate  predictions that do not match the underlying audio. \sysname\ yields consistent performance improvements on  overall WERs and rare word predictions in both in-domain and out-of-domain evaluation settings for code-mixed Hindi-English ASR tasks across four domains.

\section{Acknowledgements}
The authors
from IIT Bombay gratefully acknowledge support
from IBM Research, specifically the IBM AI Horizon Networks-IIT Bombay initiative.

\bibliographystyle{IEEE}
{\normalfont \bibliography{Bibliography}}

\end{document}

%% file: Intro.tex
\section{Introduction}
\label{sec:intro}


End-to-end RNN Transducer (RNN-T) models~\cite{Graves2012} are rapidly becoming the de facto model for streaming speech recognition systems. RNN-T models consist of two independent encoders: an acoustic encoder that processes an acoustic signal one frame at a time,
and, a language encoder consisting of an auto-regressive model that encodes previous tokens and acts as an implicit language model. Outputs from the acoustic and language encoders 
are combined and input to a joint network that makes the final RNN-T predictions. Such an architecture conveniently integrates training of the language model along with the acoustic model, and enables streaming ASR.

While RNN-T models perform competitively on various benchmarks~\cite{rao2017exploring,he2019streaming}, they struggle with rare word predictions~\cite{ravi2020improving}. For words that rarely appear in the training data, the language encoder of the RNN-T could lead the network to favour hypotheses that are unfaithful to the underlying speech. This problem is compounded when dealing with out-of-domain ASR evaluations where the test speech is drawn from domains unseen during training. For example, our baseline RNN-T system hallucinated  ``canara bank" as ``amazon" when trained on retail data but tested on banking data. More such cases of gross overriding of the acoustic input by LM hallucinations appear in Table~\ref{tab:anecdotes}.

Recent work on RNN-T models have used LM fusion techniques to improve rare word ASR predictions~\cite{sriram2017cold,peyser2020improving,ravi2020improving} but have typically relied on external resources (raw text, lexicons, etc.). In this work, we adopt a more direct approach by focusing on the RNN-T model itself and proposing model changes that aim at arresting an overconfident implicit LM from making predictions that are at odds with the underlying speech. Our proposed technique \sysname\ has the following two main features:
\begin{asparaitem}
\item We randomly mask outputs from the language encoder to the joint network, thus making the RNN-T model more robust to its spurious outputs.
\item We propose a new dynamic discounting scheme that identifies regions of interest by examining discrepancies between the acoustic and language encoders and computing the rarity of recent tokens in the predicted hypothesis. 
\end{asparaitem}

On out-of-domain evaluations, we observe statistically significant reductions in overall WER ($4\%$ relative) and large relative PER improvements of up to $14\%$ on rare words. 


%% file: relatedwork.tex
\vspace{0.5em}
\noindent \textbf{Related Work.} Prior work has handled rare word prediction and out-of-vocabulary detection~\cite{parada2010,4430159,4960493} by modeling at the subword level using hybrid LMs~\cite{bisani2005,1326093,li2017acoustictoword}.
More recent work on end-to-end models has mainly tackled the problem of out-of-domain tokens using techniques such as shallow fusion~\cite{gulcehre2015using} where hypotheses are rescored using an external LM trained on text in the target domain. Many variants of shallow fusion use some form of contextual LMs~\cite{50651,ravi2020improving,le2020deep,zhao2019shallow} that focus on fixing words appearing in specific contexts (e.g., involving named entities). A popular extension of shallow fusion, routinely employed in RNN-T models, is the density ratio technique~\cite{mcdermott2020density} that additionally discounts the scores using an external LM trained on text from the training domains. This technique has seen further improvements by discounting using an implicit LM instead of an external LM~\cite{variani2020hybrid,meng2020internal}. Both shallow fusion and the density ratio approach are inference-based techniques and hence preferred more than other techniques like deep fusion~\cite{gulcehre2015using} and cold fusion~\cite{sriram2017cold} that require retraining with the new LMs. Besides LM rescoring, other techniques addressing tail performance include the use of subword regularization~\cite{le2020deep}, using data augmentation via pronunciation dictionaries~\cite{le2020deep,zhao2019shallow,kitaoka2021dynamic} or synthetic samples~\cite{li2020developing} and finetuning the prediction network using target text~\cite{pylkkonen2021fast}. In concurrent work, \cite{zhang2022improving} tries to limit LM overtraining by combining the two encoded representations using a gate and using a regularizer to slow the training of the LM. In contrast to most of the above-mentioned prior work,
our technique fits within existing RNN-T architectures without requiring any additional parameter training or requiring external resources.

%% file: results.tex
\input{final_tables/table1}
\input{final_tables/outdomainexp}

\noindent \textbf{Datasets.} We conduct experiments on a proprietary dataset consisting of code-mixed speech (Hi-En) in a call-centre setting. This dataset contains $628$ hours of speech and is divided into four domains: Banking ($184.87$ hrs), Insurance ($165.08$ hrs), Retail ($135.87$ hrs) and Telco ($142.33$ hrs).
Domain-wise durations of train/test splis can be found in Table~\ref{tab:indomain}. 

\noindent \textbf{Architecture.} We use a baseline RNN-T model where the transcription network consists of $6$ Conformer~\cite{gulati2020conformer} encoder layers with an internal representation size of $640$ and the prediction network is a single layer LSTM with a $768$ dimensional hidden state. The conformer consists of $8$ attention heads of dim $27$ and has a kernel size of $31$ for the convolutional layers. The output of the transcription network is projected to a $256$ dim vector $h_t$. The LSTM hidden state outputs a $10$ dim embedding which is further projected to a $256$ dim vector $g_u$. The joint network consists of an additive operation between $h_t$ and $g_u$ followed by a $\tanh$ non-linearity. This is projected to the final layer of size $123$ that corresponds to the size of our character-based vocabulary $\vocab$. We use the \emph{Noam} optimizer~\cite{vaswani2017attention} for training with a learning rate of 0.0001. LM masking while training is performed with a probability of $\eta=0.2$ and weights for the implicit LM and AM are $\alpha=\beta=0.125$. The hyperparamters $\lambda$ and $\rho$ are chosen by tuning on a held-out set.

\noindent \textbf{Baselines.} Our in-domain results are compared against a vanilla RNN-T and an ILMT model trained with an implicit LM \cite{meng2021internal}. For out-of-domain experiments, we also compare with the density ratio approach~\cite{mcdermott2020density}  using external RNN-LMs (1024 units, 1 layer) trained on the source and target datasets.

\noindent \textbf{Results.} We show  WERs/PERs/CERs for both in-domain and out-of-domain settings, where test utterances are drawn from domains that are either seen or unseen during training, respectively. Along with evaluations  on the entire test set in each setting, we also separately compute PERs and CERs for rare words that appear less than $20$ times during training.%
\footnote{We do not report rare word WERs since they were $100\%$ or higher.}

Table~\ref{tab:indomain} shows in-domain results from training on speech from all domains, and four other training settings where we leave out one domain at a time. \sysname\ outperforms both RNN-T and ILMT in terms of WER reductions in almost all settings. Reductions in PER using \sysname\ for the rare words are much more significant.  Table~\ref{tab:outofdomain} shows out-of-domain results on four test domains that were each held out during training. Relative to the in-domain setting, \sysname\ outperforms all the baselines on overall WERs by a statistically significant margin (at $p < 0.001$ using the MAPSSWE test~\cite{266481}). \sysname\ gives large PER improvements on rare words for all four test domains. These PER improvements predominantly stem from fixing the errors introduced by LM hallucinations of the RNN-T baseline. Table~\ref{tab:anecdotes} provides a few examples of such  hallucinations by the RNN-T baseline where the predictions are at times completely inconsistent with the underlying acoustics (e.g., \textit{business}$\rightarrow$\textit{discount}, \textit{call}$\rightarrow$\textit{contract}, etc.). We also show examples of \sysname\ predictions that may not be identical to the ground truth but are acoustically faithful, unlike the RNN-T baseline (e.g., \textit{ikatees}$\rightarrow$\textit{iktees}, \textit{vicky}$\rightarrow$\textit{vikkee}, etc.). 

Table~\ref{tab:ablation} shows an ablation analysis of \sysname . We compare a baseline RNN-T with variants  containing $\ILM$ and $\AM$ losses (as in eqn~\eqref{eq:train_loss}) and TN masking. Discounting is particularly beneficial for rare word predictions.%
\footnote{Discounting the entire utterance with a static discount factor, instead of our adaptive scheme in \sysname, worsens overall WERs.}

 

\input{final_tables/Examples}
\input{final_tables/table4}

%% file: final_tables/table1.tex
\begin{table}[b] 
\centering
\setlength{\tabcolsep}{4pt}
\resizebox{0.5\textwidth}{!}{%
\begin{tabular}{|l|l|l|c|c|c|c|c|}
\hline
Train
(Hours) / Test (Hours) &
  
  ASR System &
  \multicolumn{1}{l|}{WER} &
  \multicolumn{1}{l|}{CER} &
  \multicolumn{1}{l|}{PER} &
  \multicolumn{1}{l|}{\begin{tabular}[c]{@{}l@{}}Rare PER\end{tabular}} &
  \multicolumn{1}{l|}{\begin{tabular}[c]{@{}l@{}}Rare CER\end{tabular}} \\ \hline \hline
\multirow{3}{*}{\begin{tabular}[c]{@{}l@{}}All-train ($477$ hrs) \\ / All-test ($75$ hrs)\end{tabular}} &

  RNNT &
  14.5 &
  13.5 &
  12.0 &
  57.7 &
  71.2 \\ \cline{2-7} 
 &
   
  ILMT &
  14.5 &
  13.4 &
  12.1 &
  53.8 &
  66.6 \\ \cline{2-7} 
 &
   
  \sysname &
  \textbf{14.1} &
  \textbf{13.1} &
  \textbf{11.8} &
  \textbf{52.2} &
  \textbf{63.9} \\ \hline \hline
  \multirow{3}{*}{\begin{tabular}[c]{@{}l@{}} Telco + Retail +  Insurance \\ ($327$ hrs) / Test ($58$ hrs)\end{tabular}} &
  RNN\_T &
  14.5 &
  13.6 &
  12.3 &
  66.9 &
  68.2 \\ \cline{2-7} 
 
   &
  ILMT &
  14.6 &
  13.4 &
  12.0 &
  69.3&
  70.5 \\ \cline{2-7} 
 &
   
  \sysname &
  \textbf{14.2} &
  \textbf{13.1} &
  \textbf{11.9} &
  \textbf{63.2} &
  \textbf{59.6} \\ \hline \hline
\multirow{3}{*}{\begin{tabular}[c]{@{}l@{}}Banking + Retail + Insurance \\ ($374$ hrs) / Test ($55$ hrs)\end{tabular}} &
  RNN\_T &
  15.2 &
  14.0 &
  12.7 &
   68.0 &
 76.6 \\ \cline{2-7} 
 &
  ILMT & 15.9&	14.5	& \textbf{12.3} &63.8&	74.2
\\ \cline{2-7} 
 &
   
  \sysname &
  \textbf{14.9} &
  \textbf{13.6} &
  \textbf{12.3} &
  \textbf{60.2} &
  \textbf{63.3} \\ \hline \hline
\multirow{3}{*}{\begin{tabular}[c]{@{}l@{}}Banking + Telco + Insurance \\ ($381$ hrs) / Test ($55$ hrs)\end{tabular}} &
  RNN\_T &
  14.2 &
  13.2 &
  12.0 & 
  65.3&
  72.7 \\ \cline{2-7} 
 
   &
  ILMT &
  14.1 &
  13.8 &
  11.8 &
64.6 &
  65.7 \\ \cline{2-7} 
 &
   
  \sysname &
  \textbf{13.8} &
  \textbf{12.9} &
  \textbf{11.6} &
  \textbf{63.1} &
  \textbf{63.8} \\ \hline \hline
\multirow{3}{*}{\begin{tabular}[c]{@{}l@{}}Banking + Telco + Retail \\ ($349$ hrs) / Test ($57$ hrs)\end{tabular}} &
  RNN\_T &
  15.0 &
  14.4 &
  13.1 &
  66.9 &
  68.2 \\ \cline{2-7} 
 &
   
  ILMT &
  15.9 &
  14.5 &
  13.2 &
69.3&
 70.5 \\ \cline{2-7} 
 &
   
  \sysname &
  \textbf{15.1} &
  \textbf{14.0} &
  \textbf{12.7} &
  \textbf{63.5} &
  \textbf{59.6} \\ \hline 
\end{tabular}
}
\caption{In-domain ASR Results.}
\label{tab:indomain}
\end{table}

%% file: final_tables/outdomainexp.tex
\begin{table}[]
\centering
\setlength{\tabcolsep}{6pt}
\resizebox{0.46\textwidth}{!}{%
\begin{tabular}{|l|l|c|c|c|c|c|}
\hline
Test Data &
  ASR-System &
  WER &
  CER &
  PER &
  \begin{tabular}[c]{@{}l@{}}Rare PER\end{tabular} &
  \begin{tabular}[c]{@{}l@{}}Rare CER\end{tabular} \\ \hline \hline
  
\multirow{4}{*}{\begin{tabular}[c]{@{}l@{}}Test-\\Banking \\(110 hrs)\end{tabular}} &
  RNN-T &
  22.4 &
  20.0 &
  18.5 &
  70.8 &
  75.9\\ \cline{2-7} 
 &
  D.R. &
  22.2 &
  19.9 &
  18.4 &
  70.8 &
  74.6 \\ \cline{2-7} 
 &
  ILMT &
 {22.4} &
  {20.0} &
 {18.5} &
 {70.9} &
  {75.9} \\ \cline{2-7} 
 &
  \sysname &
  \textbf{21.5} &
  \textbf{18.7} &
  \textbf{17.1} &
  \textbf{67.4} &
  \textbf{71.8} \\ \hline \hline
\multirow{4}{*}{\begin{tabular}[c]{@{}l@{}}
Test-\\Insurance\\(95 hrs)\end{tabular}} &
  RNN-T &
  18.4 &
  17.7 &
  16.1 &
  70.2 &
  76.7 \\ \cline{2-7} 
 &
  D.R. &
  18.1 &
  17.7 &
  16.0 &
  69.1 &
  75.6 \\ \cline{2-7} 
 &
  ILMT &
  {18.7} &
  {17.6} &
  {15.9} &
  {65.7} &
  {72.2} \\ \cline{2-7} 
 &
  \sysname &
  \textbf{17.7} &
  \textbf{16.4} &
  \textbf{14.8} &
  \textbf{60.7} &
  \textbf{67.2} \\ \hline \hline
\multirow{4}{*}{\begin{tabular}[c]{@{}l@{}} Test-\\Retail\\ (71 hrs)\end{tabular}} &
  RNN-T &
  24.7 &
  22.4 &
  20.6 &
  76.7 &
  81.3 \\ \cline{2-7} 
 &
  D.R. &
  24.6 &
  22.1 &
  20.4 &
  76.0 &
  81.3 \\ \cline{2-7} 
 &
  ILMT &
  {24.5} &
 {21.4} &
  {19.7} &
  {72.4} &
  {\textbf{77.5}} \\ \cline{2-7} 
 &
  \sysname &
  \textbf{23.7} &
  \textbf{21.2} &
  \textbf{19.5} &
  \textbf{72.2} &
  78.3 \\ \hline \hline
\multirow{4}{*}{\begin{tabular}[c]{@{}l@{}}Test-\\Telco\\ (75 hrs)\end{tabular}} &
  RNN-T &
  19.4 &
  18.6 &
  17.1 &
  68.3 &
  72.6 \\ \cline{2-7} 
 &
  D.R. &
  19.1 &
  18.2 &
  16.8 &
  70.5 &
  73.7 \\ \cline{2-7} 
 &
  ILMT &
  {19.3} &
  {18.1} &
  {16.4} &
  {\textbf{63.8}} &
  {68.9} \\ \cline{2-7} 
 &
  \sysname &
  \textbf{18.6} &
  \textbf{17.5} &
  \textbf{16.0} &
  64.3 &
  \textbf{68.6} \\ \hline 
\end{tabular}}
\caption{Out-domain ASR Results.}
\label{tab:outofdomain}
\end{table}

%% file: final_tables/Examples.tex

\begin{table}
\centering
\label{tab:my-table}
\resizebox{0.45\textwidth}{!}{
\begin{tabular}{|l|l|} 
\hline
\begin{tabular}[c]{@{}l@{}}Groundtruth\\RNN-T\\\sysname\end{tabular}           & \begin{tabular}[c]{@{}l@{}}\textit{Ki aap apanaa }\textcolor{blue}{business (\textipa{bIzn@s})} account\textit{ kab }\textcolor{blue}{op}\textcolor{blue}{en (\textipa{@Up@n})} \textit{karvana}\\\textit{ki aap apanaa} \textcolor{blue}{ discount (\textipa{dIskaUnt})} account \textit{kab } \textcolor{blue}{\textit{bund (\textipa{b@nd}) }} karvana\\ \textit{ki aap apanaa }\textcolor{blue}{business (\textipa{bIzn@s})} account \textit{kab }\textcolor{blue}{open (\textipa{@Up@n})}\textit{ }karvana\textit{~}\end{tabular}  \\ 
\hline
\hline
\begin{tabular}[c]{@{}l@{}}\textit{Groundtruth}\\\textit{ RNN-T}\\\textit{ \sysname }\end{tabular} & \begin{tabular}[c]{@{}l@{}}\textit{jisakaa pataa hain }\textcolor{blue}{plot (\textipa{pl6t})} number \textit{\textcolor{blue}{ikatees (\textipa{ikti:s})} a..}\\\textit{jisakaa pataa hai~}\textcolor{blue}{block (\textipa{bl6k} )} number \textit{ \textcolor{blue}{tees (\textipa{ti:s})} ek}\\\textit{jisakaa pataa hai }\textcolor{blue}{plot (\textipa{pl6t})} number \textcolor{blue}{\textit{iktees (\textipa{ik\textcorner \:t i:s})}}\textit{ e}..\end{tabular}                                                                                 \\ 
\hline
\hline
\begin{tabular}[c]{@{}l@{}}Groundtruth\\ RNN-T\\ \sysname\end{tabular}         & \begin{tabular}[c]{@{}l@{}}..\textit{ab aapki} \textcolor{blue}{call (\textipa{kOl} )} transfer \textit{kar dee jaayegi..}\\..\textit{ab aap}ki \textcolor{blue}{contract (\textipa{kOnt\*r \ae kt})} transfer \textit{kar dee jaayegi.}.\\..\textit{ab aapk}i \textcolor{blue}{call (\textipa{kOl} )} transfer \textit{kar dee jaayegi.}.\end{tabular}                                                                                                                                   \\ 
\hline
\hline
\begin{tabular}[b]{@{}l@{}}Groundtruth\\ RNN-T\\ \sysname\end{tabular} & \begin{tabular}[b]{@{}l@{}}..\textit{mera pooraa naam hai..} \textcolor{blue}{vicky rajak (\textipa{vikki} \textipa{\*r@d\t{Z}@k})}..\\..\textit{mera pooraa naam hai.. \textcolor{blue}{reeti raagav (\textipa{ri:ti} \textipa{ra:g@v})..}}\\..\textit{mera pooraa naam hai .. \textcolor{blue}{vikkee raaj (\textipa{vikki:} \textipa{ra:d\t{Z}})..}}\end{tabular} \\ 
\hline
\hline
\begin{tabular}[c]{@{}l@{}}Groundtruth\\ RNN-T\\ \sysname\end{tabular}         & \begin{tabular}[c]{@{}l@{}}..\textit{ye online} \textcolor{blue}{activate (\textipa{\ae ktIvejt}) }\textit{karavaa sakataa kee naheen }\\.\textit{.mujhe online} \textcolor{blue}{ network (\textipa{nEtw\*r \s{k}}) }\textit{karavaa sakataa hoon }\\..\textit{ye online} \textcolor{blue}{activate (\textipa{\ae ktIvejt})}\textit{ karavaa sakataa hoon.. }\end{tabular}                                                                                                                               \\ 
\hline
\hline
\begin{tabular}[c]{@{}l@{}}Groundtruth\\RNN-T\\\sysname\end{tabular}           & \begin{tabular}[c]{@{}l@{}}\textit{namaskaar sir main apane}~\textcolor{blue}{loan (\textipa{lown})}~\textit{ko fir se}\\ \textit{namaskaar sir main apane} \textcolor{blue}{block (\textipa{bl6k})}\textit{ ko fir se }\\\textit{namaskaar sir main apane}~\textcolor{blue}{loan  az(\textipa{lown})} \textit{ko fir se}\end{tabular}                                                                                                                                                                         \\ 
\hline
\end{tabular}
}
\caption{Anecdotes illustrating LM hallucinations. Italicized text is Romanized Hindi. Tokens of interest are in blue.}
\label{tab:anecdotes}
\end{table}

%% file: final_tables/table4.tex
\begin{table}[h]
\centering
\setlength{\tabcolsep}{6pt}
\resizebox{0.45\textwidth}{!}{%
\begin{tabular}{|l|c|c|c|c|c|}
\hline 
 &
  \multicolumn{1}{l|}{WER} &
  \multicolumn{1}{l|}{CER} &
  \multicolumn{1}{l|}{PER} &
  \multicolumn{1}{l|}{\begin{tabular}[c]{@{}l@{}}Rare PER\end{tabular}} &
  \multicolumn{1}{l|}{\begin{tabular}[c]{@{}l@{}}Rare CER\end{tabular}} \\ 
  \hline
RNN-T &
  14.5 &
  13.5 &
  12.0 &
  57.7 &
  71.2 
  \\ \hline
RNN-T + $\ILM$ + $\AM$ &
  14.5 &
  13.2 &
  11.9 &
  53.4 &
  66.5 
  \\ \hline
\begin{tabular}[c]{@{}l@{}}RNN-T+$\ILM$ +$\AM$ \\ + TN mask\end{tabular} &
  14.4 &
   \textbf{13.1} &
   \textbf{11.8} &
  53.1 & 
  65.9
  \\ \hline

  \sysname &
  \textbf{14.1} &
  \textbf{13.1} &
  \textbf{11.8} &
  \textbf{52.2} &
  \textbf{63.9} \\ \hline
  \end{tabular}
}
\caption{Ablations study to demonstrate the effect of TN masking and implicit LM/AM training in \sysname.}
\label{tab:ablation}
\end{table}